\documentclass[10pt,twocolumn,letterpaper]{article}
\usepackage{cuted}
\usepackage{capt-of}
\usepackage{etoolbox}
\usepackage{float}
\usepackage{lmodern}
\usepackage[pagenumbers]{cvpr} % To force page numbers, e.g. for an arXiv version

\usepackage{mathtools} 

\newcommand*{\Mname}{\textsc{SDiT}}

\usepackage[many]{tcolorbox}    	% for COLORED BOXES (tikz and xcolor included)
\usepackage{multicol}               % for MULTICOLUMNS

\usepackage{booktabs}
\usepackage{multirow}

\usepackage[table,dvipsnames]{xcolor}
\definecolor{main}{HTML}{283618}    % setting main color to be used
\definecolor{sub}{HTML}{FBF8CC}     % setting sub color to be used

\definecolor{mainI}{HTML}{283618}    % setting main color to be used
\definecolor{subI}{HTML}{FFCFD2}     % setting sub color to be used

\definecolor{cvprblue}{rgb}{0.21,0.49,0.74}
\usepackage[pagebackref,breaklinks,colorlinks,allcolors=cvprblue]{hyperref}

\def\paperID{*****} % *** Enter the Paper ID here
\def\confName{CVPR}
\def\confYear{2026}

\title{\Mname{}: Semantic Region-Adaptive for Diffusion Transformers}

\author{
Bowen Lin$^{1}$ \quad
Fanjiang Ye$^{2}$ \quad
Yihua Liu$^{6}$ \quad
Zhenghui Guo$^{1}$ \quad
Boyuan Zhang$^{3}$\\
Weijian Zheng$^{4}$ \quad
Yufan Xu$^{5}$ \quad
Tiancheng Xing$^{6}$ \quad
Yuke Wang$^{2}$\quad
Chengming Zhang$^{1}$\thanks{Corresponding author.}
\\[0.8em]
$^{1}$University of Houston \quad
$^{2}$Rice University \quad
$^{3}$Indiana University Bloomington\\
$^{4}$Argonne National Laboratory \quad
$^{5}$National University of Singapore \quad
$^{6}$Independent Researcher
}

\begin{document}
\twocolumn[{%
\maketitle
\vspace{-6mm}
\begin{center}
\includegraphics[width=0.86\textwidth]{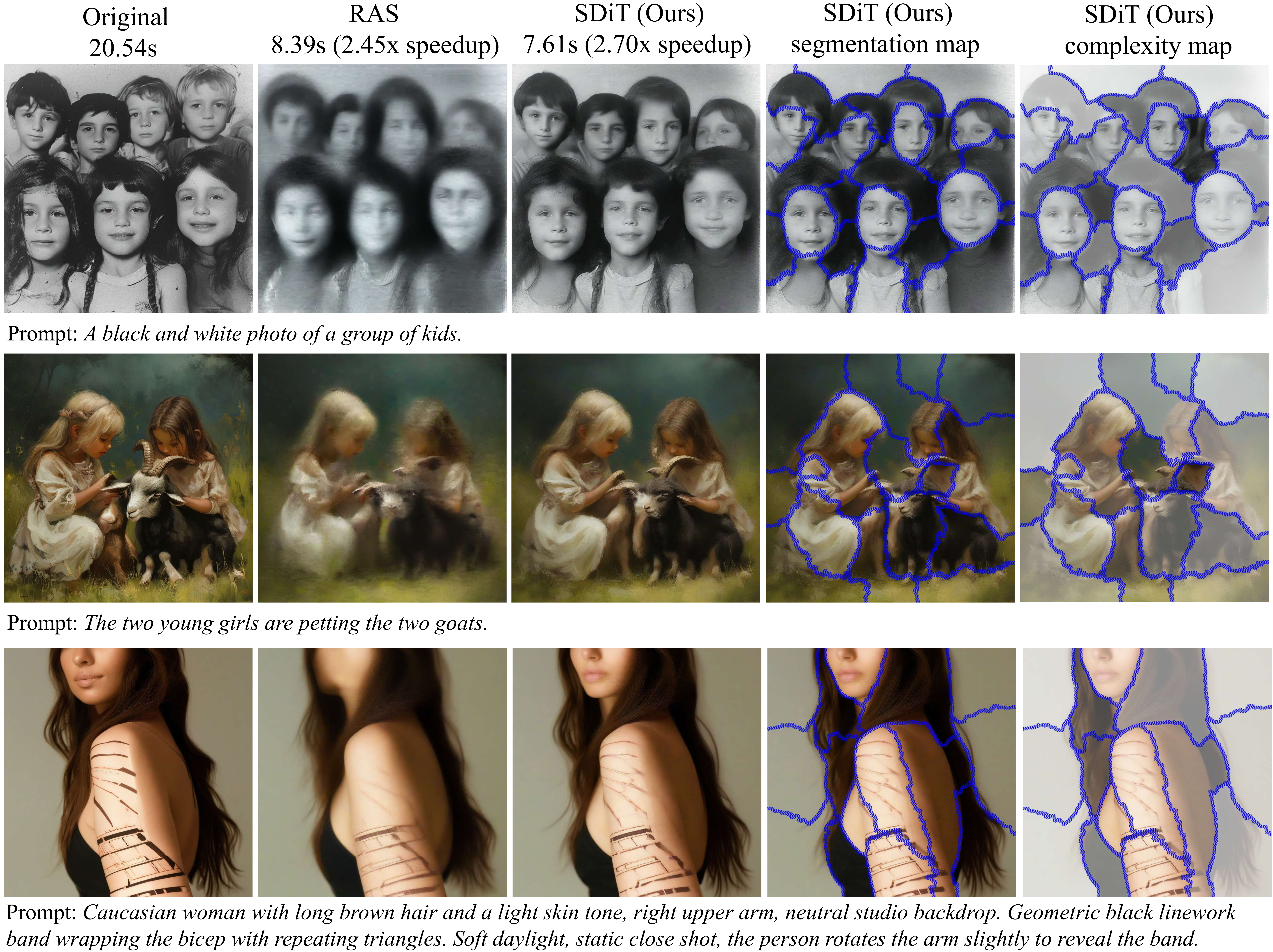}
\captionof{figure}{
\textbf{\Mname{} accelerates image generation while maintaining high fidelity.}
We introduce \Mname{}, a training-free algorithm that accelerates diffusion transformer inference by dynamically allocating computation across semantically coherent regions, achieving substantial speedup while preserving image fidelity. Under the same model and step budget, our method achieves a 2.70× speedup with sharp details and good boundary preservation. Even on the most challenging cases—such as single or multi-face generation, \Mname{} consistently preserves fine details and structural coherence. In contrast, RAS struggles to maintain pixel fidelity (e.g., blurred or collapsed facial textures) across object regions. The rightmost columns visualize \Mname{}’s segmentation and complexity maps which guide adaptive region-wise denoising. Setting: Lumina-Next with 30-step, 1024 × 1024 resolution. Latency is measured on Ada6000.}
\label{fig:intro}
\end{center}
}]

\begin{abstract}
Diffusion Transformers (DiTs) achieve state-of-the-art performance in text-to-image synthesis but remain computationally expensive due to the iterative nature of denoising and the quadratic cost of global attention. In this work, we observe that denoising dynamics are spatially non-uniform—background regions converge rapidly while edges and textured areas evolve much more actively. Building on this insight, we propose \Mname{}, a Semantic Region-Adaptive Diffusion Transformer that allocates computation according to regional complexity. \Mname{} introduces a training-free framework combining (1) semantic-aware clustering via fast Quickshift-based segmentation, (2) complexity-driven regional scheduling to selectively update informative areas, and (3) boundary-aware refinement to maintain spatial coherence. Without any model retraining or architectural modification, \Mname{} achieves up to 3.0× acceleration while preserving nearly identical perceptual and semantic quality to full-attention inference. 
% Extensive experiments on COCO and synthetic datasets demonstrate that \Mname{} not only reduces computational cost but also surpasses uniform sampling baselines under matched budgets, offering a principled path toward efficient and adaptive diffusion generation.
\end{abstract}    
\section{Introduction}
\label{sec:intro}

Diffusion models~\cite{ho2020denoising} have achieved remarkable success in generative modeling, producing images with high fidelity and semantic coherence. Beyond static image synthesis~\cite{peebles2023scalable}, they have shown strong capability across diverse visual tasks, including image editing~\cite{meng2022sdeditguidedimagesynthesis, hertz2022prompttopromptimageeditingcross}, inpainting~\cite{lugmayr2022repaintinpaintingusingdenoising, saharia2022paletteimagetoimagediffusionmodels}, painting~\cite{nichol2022glidephotorealisticimagegeneration, rombach2022highresolutionimagesynthesislatent}, video synthesis~\cite{ho2022videodiffusionmodels}, and 3D generation~\cite{poole2022dreamfusiontextto3dusing2d, lin2023magic3dhighresolutiontextto3dcontent}. Among these, Diffusion Transformers (DiTs)~\cite{peebles2023scalable} represent the state of the art by combining the expressive power of attention~\cite{vaswani2017attention} with diffusion-based denoising. However, the quadratic $O(n^2)$ attention cost and iterative sampling make DiTs computationally prohibitive for high-resolution or real-time generation. Achieving practical deployment thus requires efficient inference acceleration without retraining.

Recent training-free methods aim to accelerate diffusion inference with pre-trained models.
DPM-Solver~\cite{lu2022dpm} reformulates the reverse process as an ODE and applies high-order solvers to reduce sampling steps. DistriFusion~\cite{li2024distrifusion} improves throughput via patch-level parallelism, while $\Delta$-DiT~\cite{chen2024delta} exploits block-wise redundancy through delta-based caching. Although effective, these methods mainly reduce temporal or hardware redundancy and still process the entire feature map uniformly, ignoring spatial variation.

In practice, different image regions exhibit distinct denoising dynamics: detailed areas such as textures and boundaries evolve rapidly, while smooth backgrounds remain stable~\cite{kim2024radregionawarediffusionmodels, rombach2022highresolutionimagesynthesislatent}. Allocating equal computation to all regions, therefore wastes considerable resources. Region-Adaptive Sampling (RAS)~\cite{liu2025region} partially addresses this by assigning higher update frequency to bright color regions, but its intensity-based policy lacks semantic awareness, often leading to blurred or inconsistent results.

\begin{figure}[h] 
  \begin{center}
\includegraphics[width=\linewidth]{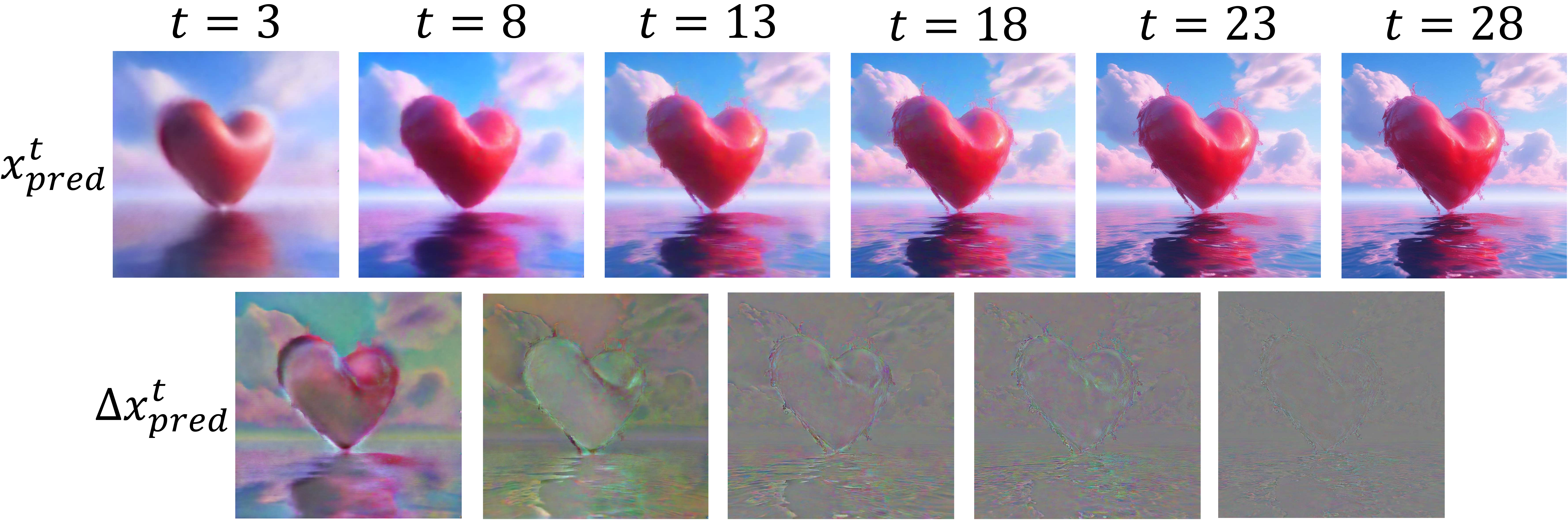}
  \end{center}
 \vspace{-4mm}
  \caption{Visualization of the diffusion process}
  \label{fig: diffusion process}
 \vspace{-4mm}
\end{figure}

To address these limitations, we propose \Mname{}, a semantic region-adaptive diffusion framework that dynamically allocates denoising computation according to semantic structure and local complexity, enabling spatially adaptive yet globally coherent generation without retraining. 

To further investigate this idea, we visualize the diffusion process at different sampling steps, as shown in Figure~\ref{fig: diffusion process}. Based on the predicted final states $x^t_{\text{pred}}$, we observe that as the timestep progresses, the information gain $\Delta x^t_{\text{pred}}$ is primarily concentrated in textured and boundary regions. A closer examination of denoising trajectories reveals non-uniform spatial evolution: background regions exhibit stable, low-complexity dynamics, while textured and boundary areas change rapidly. However, updating only high-complexity regions leads to degraded coherence, as local attention relies on contextual continuity. Low-complexity areas, though visually simple, often provide crucial cues for adjacent high-complexity zones. 

Building on these insights, \Mname{} integrates semantic understanding with adaptive regional computation to focus resources where they matter most. It features three key components: 
(1) an efficient clustering-based semantic segmentation that partitions the image into contiguous, semantically coherent regions for joint refinement and consistent global attention; 
(2) a region-adaptive complexity estimation mechanism that dynamically reallocates computation according to denoising progress, emphasizing high-frequency regions while reducing redundant updates in smooth areas 
(3) a boundary-aware enhancement module that reinforces object edges and high-frequency details to mitigate potential artifacts. 

Together, these components form a unified, zero-shot region-adaptive framework that achieves substantial acceleration while preserving texture fidelity, structural coherence, and semantic consistency, all without retraining or modifying any DiT parameters.

To validate our approach, we apply \Mname{} on a state-of-the-art text-to-image generation DiT models Lumina-Next T2I~\cite{zhuo2024lumina}. Extensive experiments demonstrate clear improvements: with the same computational budget, our approach produces notably higher generation quality, with FID score improved by 14\% using 8\% less computation and achieves comparable fidelity while achieving up to 3.0× end-to-end speedup with less than 20\% of FID score drop. We summarize our key contributions as follows: 
\begin{itemize}
    \item We conduct an in-depth analysis of the regional correlation of diffusion process, revealing that denoising should be based on semantic segmentation and dynamics.
    \item We propose \Mname{}, a training-free and efficient DiT inference system that dynamically updates regions based on semantic segmentation, achieving high-quality and efficient generation.
    \item \Mname{} delivers a significant speedup up to 3.0x while maintaining good image generation quality, paving the way for efficient image generation.
\end{itemize}

% By integrating semantic-segmentation guidance, region adaptivity, and noise momentum, our framework establishes a new paradigm for training-free, structure-preserving acceleration of diffusion transformers, bridging the gap between academic diffusion research and real-time, deployable generative systems.

\section{Related Work}
\label{sec:related}

\subsection{Latent Diffusion Transformer Models}

\begin{figure}[t] 
  \begin{center}
\includegraphics[width=\linewidth]{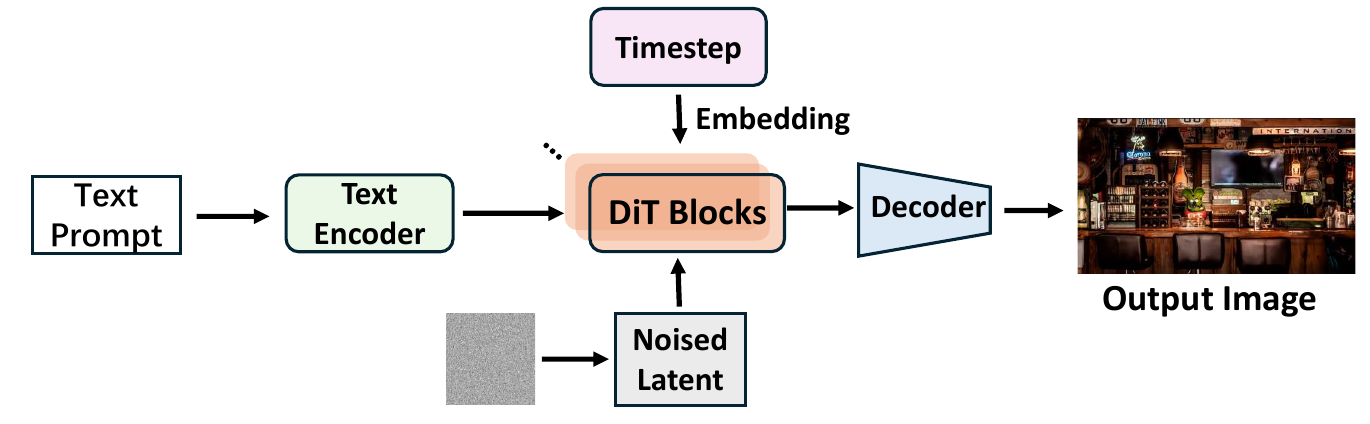}
  \end{center}
 \vspace{-4mm}
  \caption{Overview of Diffusion Transformer Structure.}
  \label{fig: DiT Structure}
 \vspace{-4mm}
\end{figure}

Latent Diffusion Models (LDMs)~\cite{rombach2022highresolutionimagesynthesislatent, ho2020denoisingdiffusionprobabilisticmodels, song2022denoisingdiffusionimplicitmodels} have become a leading paradigm for high-fidelity image synthesis by performing diffusion in a compact latent space. An encoder $\mathcal{E}$ first projects an input $x$ to a latent code $z_0=\mathcal{E}(x)$. The model then runs a two-phase stochastic process: a forward (noising) phase and a reverse (denoising) phase. In the forward phase, Gaussian noise is progressively injected into $z_0$ across $T$ timesteps according to a predetermined variance schedule $\{\beta_t\}_{t=1}^T$, yielding a latent that approaches the standard normal distribution $\mathcal{N}(0,I)$, as described in Eq.~\ref{equation: forward diffusion}. 
\vspace{-5pt}
\begin{equation} 
\label{equation: forward diffusion}
q(\mathbf{z}_t \mid \mathbf{z}_{t-1}) = \mathcal{N} \left( \mathbf{z}_t; \sqrt{1 - \beta_t} \, \mathbf{z}_{t-1}, \beta_t \mathbf{I} \right).
\end{equation}
In the reverse process, a learnable noise prediction network is utilized to reconstruct the original data $z_o$ by iteratively denoising from the noised sample $z_T$, as outlined in Equation~\ref{equation: reverse process of diffusion}, where $\mu_\theta(\mathbf{z}_t, t)$ and $\Sigma_\theta(\mathbf{z}_t, t)$ are the predicted mean and variance of the Gaussian distribution. 
\vspace{-2mm}
\begin{equation} 
\label{equation: reverse process of diffusion}
    p_\theta(\mathbf{z}_{t-1} \mid \mathbf{z}_t) = \mathcal{N} \left( \mathbf{z}_{t-1}; \mu_\theta(\mathbf{z}_t, t), \Sigma_\theta(\mathbf{z}_t, t) \right).
\end{equation}
% \vspace{-2mm}
Among various diffusion backbones, the Diffusion Transformer (DiT)~\cite{peebles2023scalablediffusionmodelstransformers, chen2023pixartalphafasttrainingdiffusion, esser2024scaling} has become particularly prominent due to its strong modeling capacity and scalability. Rather than preserving the original spatial resolution for pooling as in earlier convolutional U-Nets~\cite{rombach2022highresolutionimagesynthesislatent}, DiT operates without this constraint, making it possible to mine spatial redundancy during the diffusion process. In a typical DiT-based image generation pipeline inference as shown in Figure~\ref{fig: DiT Structure}, we sample Gaussian noise in the VAE latent space and feed it, together with timestamp embeddings and text features, into the DiT model, and after iterative denoising, the refined latent is decoded by the VAE to produce the final image. This design decouples computation from the fixed image grid: once positional embeddings are added, tokens can be addressed individually, enabling adaptive selection of salient tokens at each step (Sec.~\ref {sec:intro}) and substantially improving diffusion efficiency ~\cite{liu2025regionadaptivesamplingdiffusiontransformers}.

\subsection{Efficient Diffusion Models}
Due to the high cost of inference in diffusion models, it's necessary to apply optimizations to improve the generation efficiency, commonly including both training-free and non-training-free methods.  
\vspace{-4mm}
\paragraph{Non-training-free methods.} Many non-training-free methods aim to reduce the sampling steps. For example, the progressive distillation~\cite{yin2024improved, yin2024one} aims to train simple and few-step models. Another method, like consistency models~\cite{song2023consistencymodels}, which learns a time-consistent mapping so that predictions at different noise levels agree, enabling one- or few-step generation via consistency training. What's more, the rectified flow~\cite{albergo2023buildingnormalizingflowsstochastic, lipman2023flowmatchinggenerativemodeling}, which learns a vector field whose ODE follows near-straight trajectories from a base Gaussian to the data distribution, shortening transport paths and reducing the number of sampling steps.
\vspace{-6mm}
\paragraph{Training-free methods.} There also exist training-free approaches that reduce the number of sampling steps and the computation within each step. Methods such as DPM-solver~\cite{Lu_2025, lu2022dpm} refine the underlying ODE trajectories to enable fast integration. In parallel, caching has emerged as a practical way to lower per-step cost by reusing intermediate activations: early studies primarily targeted U-Net backbones~\cite{295597, ma2023deepcacheacceleratingdiffusionmodels}, while later efforts for DiTs explored output reuse, block skipping, and residual reuse~\cite{lv2025fastercachetrainingfreevideodiffusion, selvaraju2024forafastforwardcachingdiffusion}. However, these techniques typically treat all image regions uniformly, overlooking spatially varying complexity. Recent work therefore allocates computation adaptively across regions~\cite{liu2025region, jeong2025upsamplemattersregionadaptivelatent, ye2025supergenefficientultrahighresolutionvideo}, yielding substantial gains in efficiency.

% \paragraph{Diffusion model compression.}
% \paragraph{Efficient system implementation.}

%%%%%%%%%%%%%%%%%%%%%%%%%%%%%%%%%%%%%%%%%%
\section{Motivation and Analysis}

\vspace{-4mm}
\begin{figure}[!h] 
  \begin{center}
\includegraphics[width=.9\linewidth]{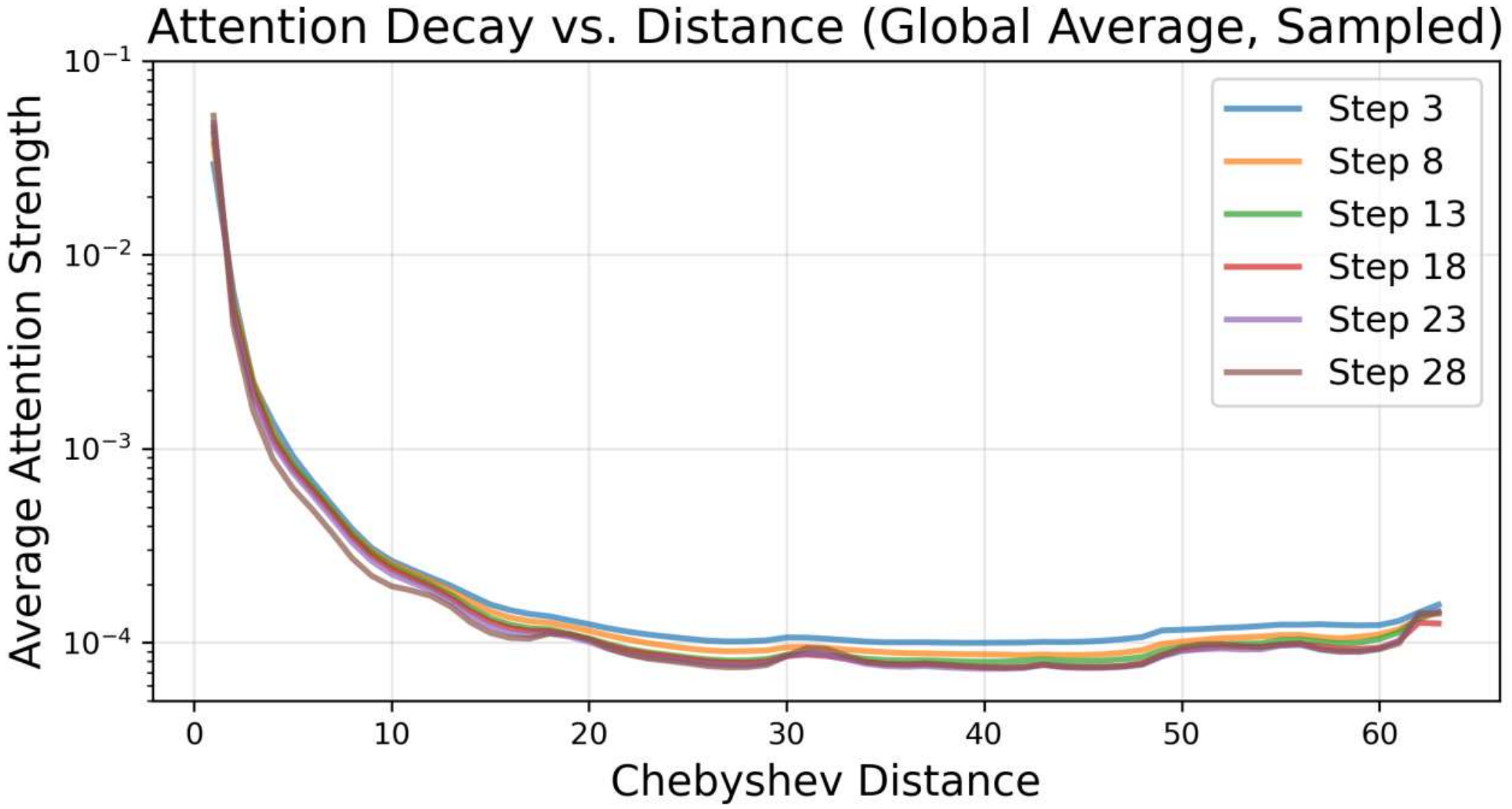}
  \end{center}
  \vspace{-4mm}
  \caption{Average attention strength as a function of Chebyshev distance across denoising steps.}
  \label{fig: attention_decay}
  \vspace{-3mm}
\end{figure}

\begin{figure}[!h] 
  \begin{center}
\includegraphics[width=\linewidth]{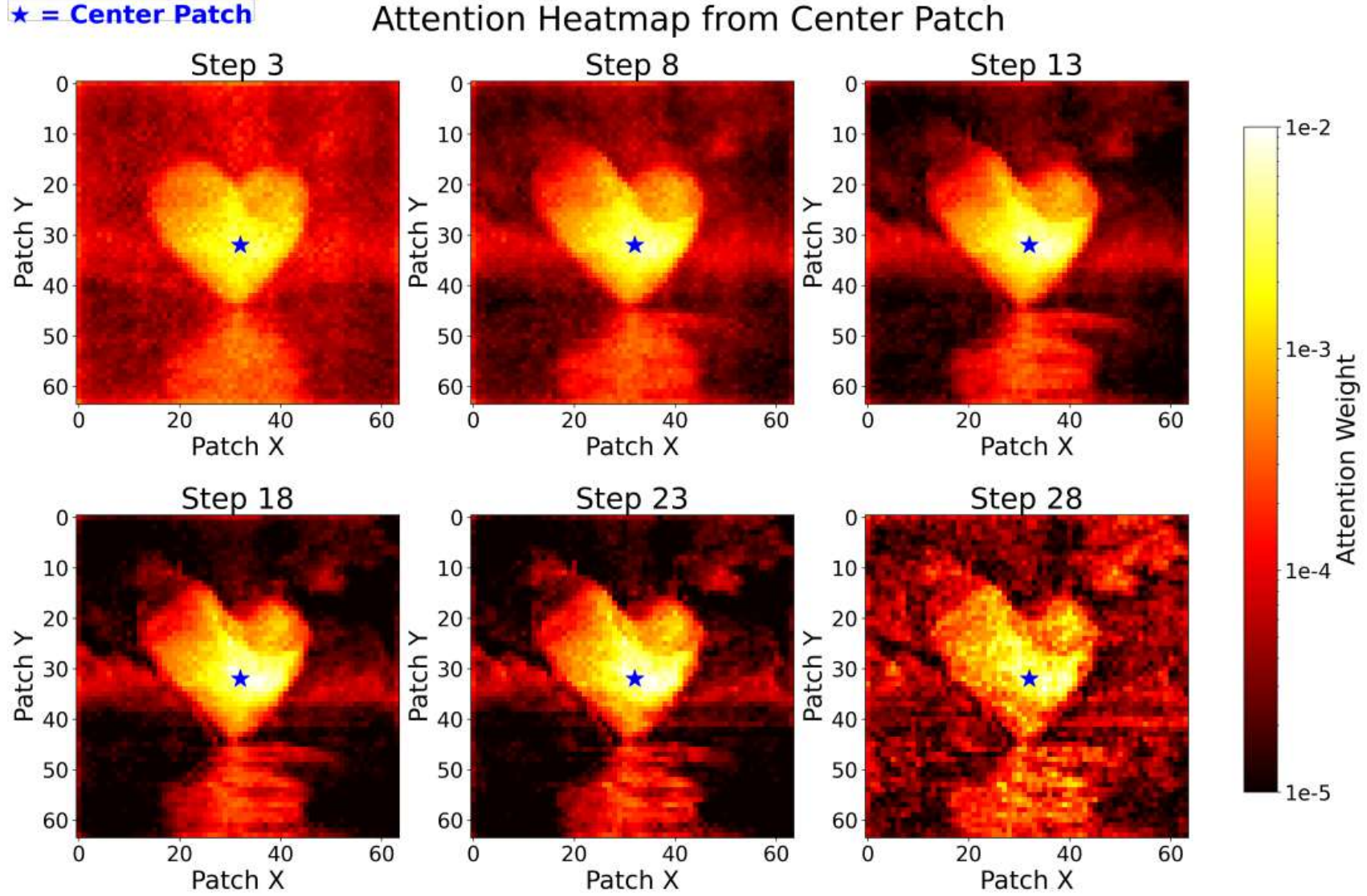}
  \end{center}
 \vspace{-3mm}
  \caption{Attention heatmaps from the center patch (marked by a blue star) across denoising steps.} 
  \label{fig: attention_heatmap}
 \vspace{-3mm}
\end{figure}

Diffusion Transformers suffer from high computational costs because each denoising step requires performing global self-attention over the entire latent feature map. A natural way to reduce this cost is to introduce \textbf{spatial sparsity}—that is, updating only a subset of regions at each timestep. However, how to divide the latent into meaningful regions while maintaining semantic coherence and attention consistency remains a key challenge for efficient sparse diffusion.

\subsection{Attention Observation}

As shown in Figure~\ref{fig: attention_decay} and Figure~\ref{fig: attention_heatmap}, the attention map from a representative center patch reveals a clear local preference: the highest attention weights are concentrated around nearby patches, while distant regions contribute minimally. This confirms that DiT’s attention mechanism primarily captures short-range spatial dependencies, reflecting an inherent locality bias in diffusion transformers.

However, a closer inspection of the heatmaps uncovers a more intriguing phenomenon: the attended regions are not randomly distributed but instead form coherent semantic clusters that align with the underlying object structure. For instance, in the example figure, the high-attention area naturally outlines the heart-shaped object and its reflection, suggesting that the model implicitly organizes attention according to semantic and structural boundaries. In other words, attention tends to follow the shape of meaningful regions rather than purely relying on Euclidean proximity.

This observation indicates that while the transformer inherently focuses on local neighborhoods, its attention also exhibits semantic grouping behavior—a property that can be leveraged to design more adaptive region-based acceleration strategies. Instead of treating all square patches equally, we can utilize this semantic prior to partition the latent space into content-aware regions that better respect object boundaries and visual coherence.

% \subsection{Motivation for Semantic Region-Adaptive Division}
% To address these issues, we propose a \textbf{semantic region-adaptive division} framework that jointly considers denoising dynamics and spatial continuity. This design yields regions that are both semantically coherent and spatially contiguous. Building upon this, we further introduce a dynamic scheduling strategy that enables true spatially sparse acceleration.

\subsection{Limitations of Existing Region Division}

A straightforward idea is to partition the latent variable into multiple regions and assign different computational budgets to each. Three simple strategies are commonly used:

\noindent \textit{K-Means clustering}~\cite{macqueen1967multivariate}: grouping pixels based on feature similarity.

\noindent \textit{Uniform patch partition}~\cite{dosovitskiy2020image}: dividing the image into fixed-size blocks.

\noindent \textit{Segmentation-model division}: using a pretrained semantic segmentation model to obtain region masks.

% \begin{enumerate}
%     \item \textbf{K-Means clustering}~\cite{macqueen1967multivariate} — grouping pixels based on feature similarity;
%     \item \textbf{Uniform patch partition}~\cite{dosovitskiy2020image} — dividing the image into fixed-size blocks;
%     \item \textbf{Segmentation-model division} — using a pretrained semantic segmentation model to obtain region masks.
% \end{enumerate}
\noindent However, all of these strategies have inherent limitations.

\noindent (1) K-Means clustering groups pixels solely based on feature similarity, completely ignoring spatial adjacency. As a result, distant pixels from the same object may be assigned to different clusters, while foreground and background pixels may be mixed due to similar colors. The resulting regions are highly fragmented, spatially disconnected, and poorly aligned with semantic structures.

\noindent (2) Uniform patch partition maintains spatial continuity but disregards structural boundaries. A single patch may contain both smooth background and sharp edges, while adjacent patches belonging to the same object are treated independently. This disrupts attention dependencies across regions, leading to inconsistent updates, broken contours, and texture artifacts.

\noindent (3) Segmentation-model-based division uses pretrained models (e.g., SAM 2~\cite{ravi2024sam2}) to derive semantic masks, but the segmentation results often yield highly unbalanced region sizes, and require expensive inference with extra memory overhead, which hinders its use for iterative adaptive scheduling during diffusion.

\begin{figure}[t] 
  \begin{center}
\includegraphics[width=\linewidth]{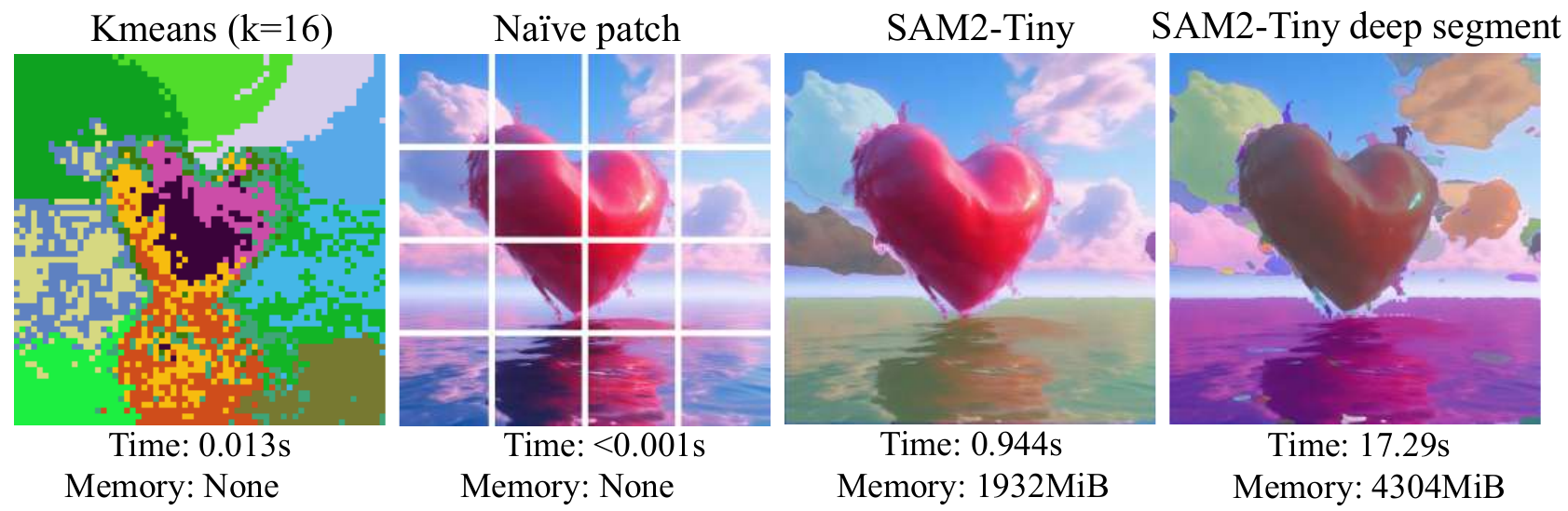}
  \end{center}
\vspace{-3mm}
  \caption{Comparison of different region segmenting methods}
  \label{fig: Method Comparison}
 \vspace{-4mm}
\end{figure}

Figure~\ref{fig: Method Comparison} visually compares these three strategies. As shown, none of them simultaneously achieves semantic integrity, spatial continuity, and computational efficiency.

\section{Methodology}

\begin{figure*}[t]
    \centering
    \includegraphics[width=0.9\textwidth, trim=10 5 10 5, clip]{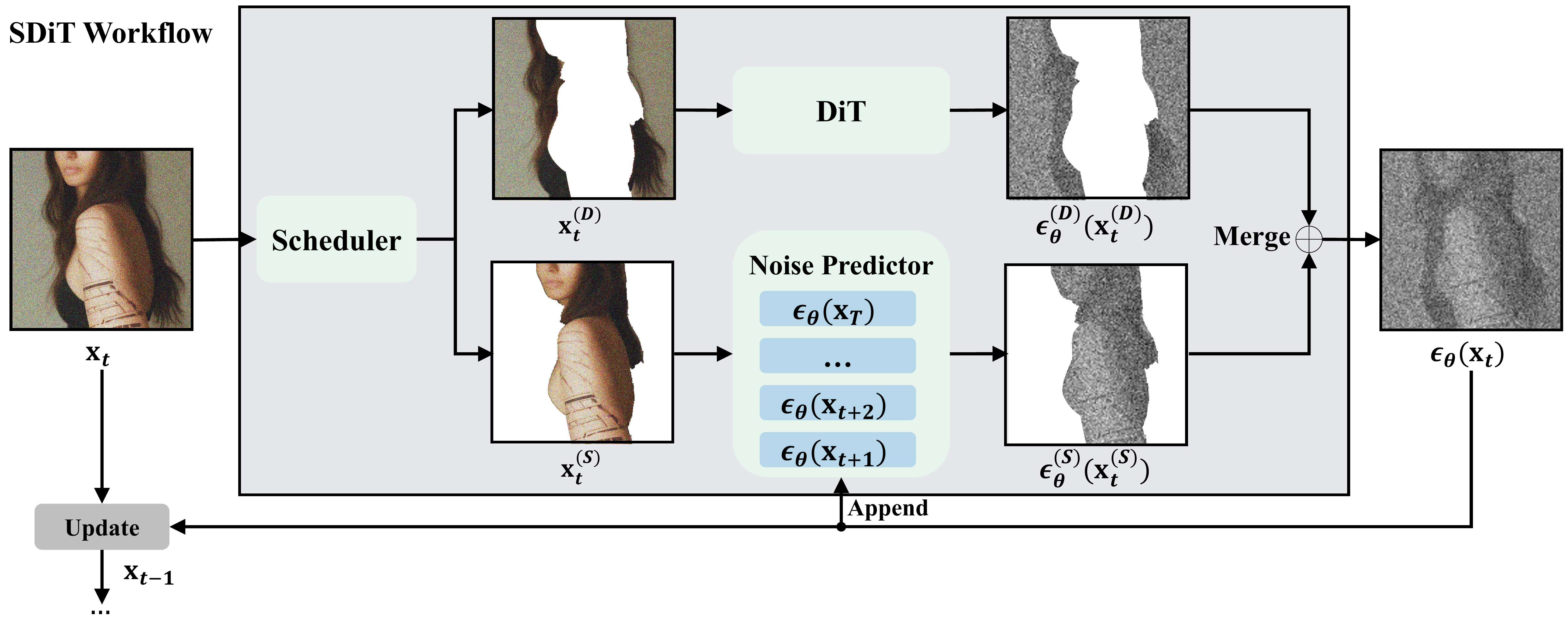}
    \vspace{-4mm}
    \caption{
        Dynamic region scheduling in \Mname{}.
        The input latent is first processed by a fast semantic segmentation module, where each region is treated as an independent unit for subsequent updates. The complexity estimator then computes the denoising complexity of each region, and the scheduler selects the top-k most complex regions for active denoising. To maintain spatial continuity and enhance high-frequency boundaries, boundary dilation is applied to the selected regions before merging. Finally, the latent space is divided into two types of regions—actively updated regions and momentum-based cached regions—achieving adaptive region-level denoising scheduling. 
    }
    \vspace{-2mm}
    \label{fig:SDiT}
\end{figure*}

\subsection{Intelligent Semantic-aware Division}

\begin{figure}[t!]
    \centering
    \includegraphics[width=.95\columnwidth, trim=10 5 10 5, clip]{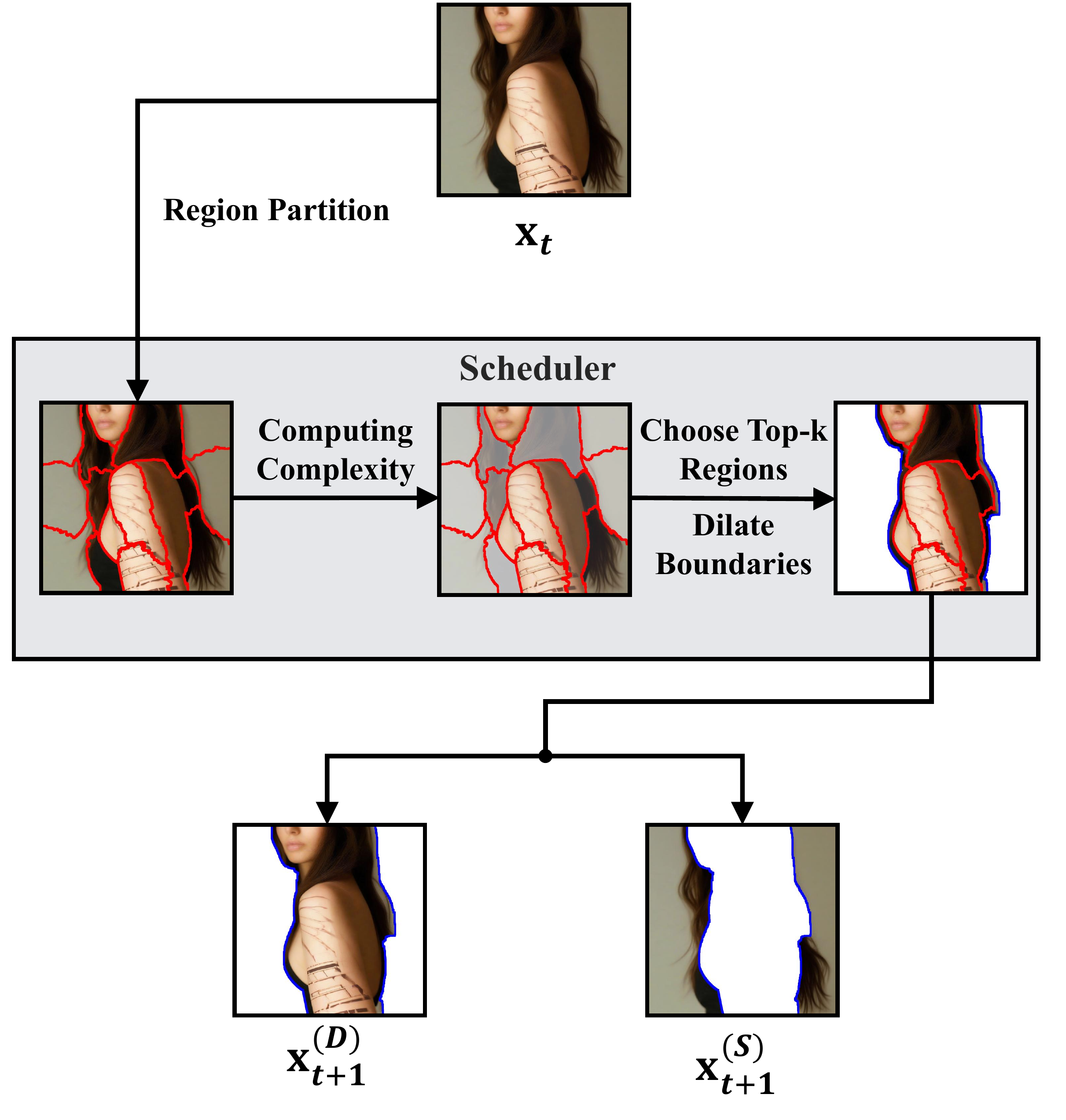}
    \vspace{-4mm}
    \caption{
        Overview of the \Mname{} Scheduler.
        The input latent is first processed by a fast semantic segmentation module, where each region is treated as an independent unit for subsequent updates. The complexity estimator then computes the denoising complexity of each region, and the scheduler selects the top-k most complex regions. To maintain spatial continuity and enhance high-frequency boundaries, boundary dilation is applied to the selected regions before performing active denoising.
    }
    \vspace{-6mm}
    \label{fig:scheduler}
\end{figure}

As illustrated in Figure~\ref{fig:SDiT}, given the latent variable $x_t$ at timestep $t$, our goal is to divide it into semantically coherent regions, enabling different regions to be updated with distinct computational budgets.

To this end, we propose a \textit{Quickshift-based dynamic clustering} method that integrates both denoising dynamics and spatial continuity. Specifically, we extract two latent features from the denoising network $\epsilon_\theta$: the predicted clean image $x_{\text{pred}}^t$ and the predicted noise $\epsilon_\theta(x_t)$. For each spatial location $i$, we construct a feature descriptor:
\begin{equation}
    f_i^t = \big[\, x_{\text{pred},i}^t \ \Vert \ \epsilon_\theta(x_t)_i \,\big].
\end{equation}

Instead of K-Means, we apply the density-based Quickshift clustering algorithm~\cite{vedaldi2008quick,jiang2018quickshift++} with approximate nearest neighbor (ANN) acceleration~\cite{malkov2018efficient}. Quickshift naturally enforces spatial connectivity and does not require a predefined number of clusters $K$, making it ideal for dynamically evolving semantics in diffusion. With ANN acceleration, the overall complexity is reduced from $O(N^2)$ to $O(N \log N)$, and the clustering stage becomes lightweight—each clustering operation takes no more than 20 ms and accounts for less than 5\% of the total inference time.

The latent space is thus partitioned into $K$ regions:
\begin{equation}
    \{ M_1, M_2, \dots, M_K \}, \quad
    M_k = \{\, i \mid \text{Quickshift}(f_i^t) = k \,\}.
\end{equation}

Here, $i$ denotes a spatial position (i.e., pixel or patch index) in the latent feature map at timestep $t$, and $\text{Quickshift}(\cdot)$ represents the clustering operator that assigns each position $i$ to one of $K$ clusters based on its local density in the feature space. 
Each cluster $M_k$ therefore corresponds to a spatially contiguous and semantically coherent region in the latent domain.

The feature descriptor combines two complementary components: 
(1) the predicted clean latent $x_{\text{pred}}^t$ encoding semantic content, and 
(2) the noise prediction $\epsilon_\theta(x_t)$ encoding denoising dynamics. 
This joint representation ensures that clustering reflects both spatial semantics and temporal denoising behavior.

Each region $M_k$ serves as the basic scheduling unit for selective denoising, re-computed each step to follow evolving semantics while preserving global context and temporal coherence.

\subsection{Efficient Region-aware Scheduling}

Figure~\ref{fig:scheduler} shows the workflow of \Mname{} scheduler. After obtaining semantic regions ${M_{k=1,\cdots,K}}$, the key challenge becomes determining which regions to update, when to update them, and how to update them without breaking spatial coherence. We propose an efficient regional scheduling strategy composed of three components: complexity-based selection, boundary-aware refinement, and velocity-space extrapolation.
\vspace{-4mm}
\paragraph{Complexity-Based Region Selection.}
For each spatial location $i$, we define a lightweight complexity score that reflects the local denoising difficulty:
\begin{align}
C_i^t = \frac{1}{C}\sum_{c=1}^C \big[
&\, \underbrace{\alpha \sqrt{(\partial_x x_{\text{pred},i,c}^t)^2 + (\partial_y x_{\text{pred},i,c}^t)^2}}_{\text{edge / texture}} \nonumber\\
&+ \underbrace{\gamma \, \lvert \Delta x_{\text{pred},i,c}^t \rvert}_{\text{fine details}}
+ \underbrace{\beta \, \lvert x_{i,c}^t - x_{\text{pred},i,c}^t \rvert}_{\text{residual noise}}
\big]
\end{align}

Here, $x_t$ and $x_{\text{pred}}^t$ denote the noisy and predicted clean latents, $i$ and $c$ are spatial and channel indices, $\partial_x, \partial_y$ are spatial gradients, and $\Delta$ is the Laplacian operator. $\alpha, \gamma, \beta$ control the weighting of edges, fine structures, and residual noise.

Instead of averaging complexity over all pixels in a region—which would dilute sharp structures with smooth areas—we compute the mean of the top $q\%$ complex pixels within each region:
\begin{equation}
    \bar{C}_k^t = \frac{1}{|\text{Top}_q(M_k)|} 
    \sum_{i \in \text{Top}_q(M_k)} C_i^t
\end{equation}

Regions are ranked by $\bar{C}_k^t$, and only the top $p(t)$ regions are selected for denoising at timestep $t$. The rest reuse cached predictions from the previous step.

\begin{figure}[t] 
  \begin{center}
\includegraphics[width=0.95\linewidth]{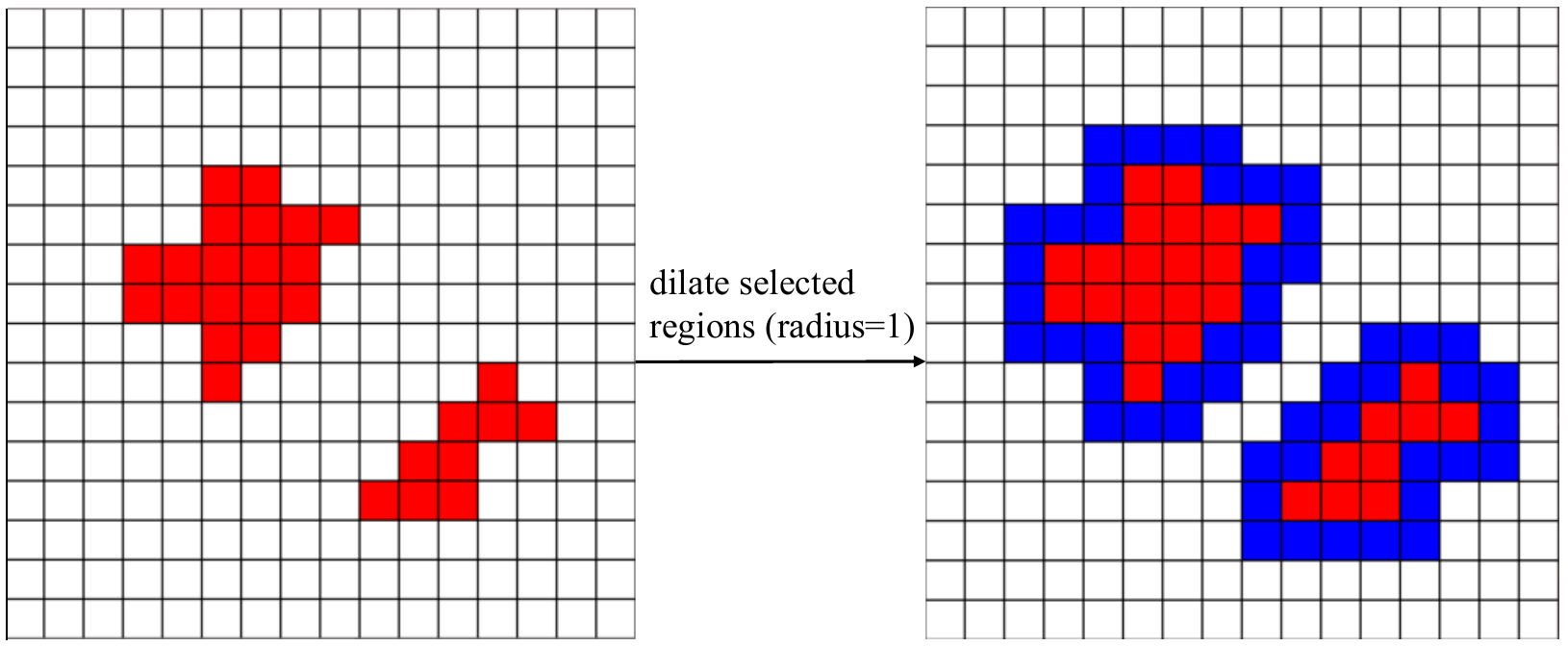}
  \end{center}
 \vspace{-3mm}
  \caption{Process of dilating boundary}
  \label{fig: boundary}
 \vspace{-6mm}
\end{figure}

% \vspace{-4mm}
\paragraph{Boundary-aware Refinement.}
Updating only selected regions may cause boundary artifacts because attention becomes asymmetric across region borders. As shown in Figure~\ref{fig: boundary}, to ensure consistent updates, we apply an 8-connected dilation to each selected region:
\begin{equation}
M_k^{\text{ext}} = \text{Dilate}(M_k, r = 1)
\end{equation}
and define the full update set as:
\begin{equation}
\Omega_t = \bigcup_{k \in \text{Top-}p(t)} M_k^{\text{ext}}
\end{equation}
This boundary expansion ensures both sides are updated, preserving contour continuity and semantic consistency.
\vspace{-4mm}
\paragraph{Non-Uniform Velocity Space Extrapolation.}
Low complexity regions change more gradually during denoising,
showing greater temporal stability and more reliable predictability than highly dynamic areas.
Taking advantage of this property, we model their latent evolution and predict the next-step noise from previous timesteps, which greatly reduces computational cost.
However, the diffusion schedule $\{\sigma_t\}$ is non-uniform, 
causing the scale of $\epsilon_\theta(x_t, t)$ to vary with the step interval. 
To achieve consistent modeling across irregular steps, 
we perform extrapolation in the \emph{velocity space}, defined as
\begin{equation}
    v_t = \frac{\epsilon_\theta(x_t, t)}{\sigma_t - \sigma_{t-1}},
\end{equation}
where $\sigma_t$ denotes the noise standard deviation at step $t$. 
Given observed pairs $\{(\sigma_i, v_i)\}_{i=0}^{t}$, 
we estimate $v_{t+1}$ via Newton interpolation with recursively defined divided differences:
\begin{equation}
\begin{aligned}
v[\sigma_i] &= v_i, \\
v[\sigma_i,\sigma_{i+1}] &= \frac{v_{i+1}-v_i}{\sigma_{i+1}-\sigma_i}, \\
v[\sigma_i,\dots,\sigma_{i+k}] &= 
\frac{v[\sigma_{i+1},\dots,\sigma_{i+k}] - v[\sigma_i,\dots,\sigma_{i+k-1}]}{\sigma_{i+k}-\sigma_i}.
\end{aligned}
\end{equation}

\noindent
We define the Newton basis term as
\vspace{-2mm}
\begin{equation}
    \alpha_k(\sigma_{t+1}) := \prod_{j=0}^{k-1}(\sigma_{t+1}-\sigma_{t-j}), 
    \quad \alpha_0 \equiv 1,
\end{equation}
and introduce a $k$-dependent decay weight to attenuate higher-order terms over large temporal gaps:
\begin{equation}
    w_k = \exp[-\lambda(\sigma_t - \sigma_{t-k})].
\end{equation}

\noindent
The extrapolated velocity is then given by
\begin{equation}
    v_{t+1} \approx 
    \sum_{k=0}^{n} 
    w_k\,\alpha_k(\sigma_{t+1})\,v[\sigma_{t-k},\dots,\sigma_t].
\end{equation}

\noindent
Finally, we convert the result back to the noise space as
\begin{equation}
    \hat{\epsilon}_{t+1} = v_{t+1} \cdot (\sigma_{t+1} - \sigma_t),
\end{equation}
which provides temporally consistent and scale-normalized noise estimates 
for stable or skipped regions under non-uniform schedules.

\subsection{Other Quality Optimizations}

\paragraph{Dynamic Sampling Ratio.}
We divide the denoising process into three perceptual phases—(1) structure formation, (2) stable refinement, and (3) detail restoration—and design a cosine-based Structure–Stable–Detail (SSD) scheduling to match these dynamics:
\begin{equation}  \small
p(t) =
 \begin{cases}
 p_{\max}, & \frac{t}{T} < \tau_1, \\
 p_{\min} + \frac{p_{\max} - p_{\min}}{2} \left( 1 + \cos \left( \pi \cdot \frac{\frac{t}{T} - \tau_1}{\tau_2 - \tau_1} \right) \right) & \tau_1 \le \frac{t}{T} \le \tau_2, \\
 p_{\max}, & \frac{t}{T} > \tau_2, \end{cases}
\end{equation}
where $\tau_1 \approx 0.1$ and $\tau_2 \approx 0.95$.

\paragraph{Global Refresh Steps.}
To prevent long-term drift in sparse updates, we insert several full-image denoising steps at critical moments:
(1) Warm-up: early steps where clustering is unstable;
(2) Cool-down: late steps for recovering fine details;
(3) Error reset: when cached predictions diverge significantly.

These steps incur negligible overhead, substantially enhancing temporal stability and perceptual consistency.

\section{Experiment}    

\subsection{Setup}
\paragraph{Models.} We evaluate our method on a state-of-the-art, open-sourced diffusion transformer models based on the flow matching paradigm: Lumina-Next T2I. This model generates images in latent space and employs transformer-based architectures with cross-frame self-attention.
\vspace{-4mm}
\paragraph{Metrics.} We evaluate both efficiency and generation quality of accelerated sampling. For fidelity, we report Fréchet Inception Distance~\cite{heusel2017gans} (FID, lower is better) computed against ground-truth images as well as the original full-step diffusion outputs. To measure reconstruction consistency with the original model, we use PSNR and SSIM~\cite{wang2004image} (higher is better) and LPIPS~\cite{zhang2018unreasonable} (lower is better), which together capture pixel-level accuracy, structural similarity, and perceptual alignment. To measure text-image alignment, we use CLIP score~\cite{hessel2021clipscore}. Efficiency is assessed using latency per image (s/image) and the corresponding speedup ratio relative to full sampling.
\vspace{-4mm}
\paragraph{Datasets.} We evaluate our method on two complementary text–image datasets to cover both real-world and synthetic generation scenarios. From MS-COCO 2017 dataset~\cite{lin2014microsoft}, we sample 5,000 caption–image pairs and use the captions to generate images at 1024×1024 resolution, testing semantic alignment and realism in natural scenes. From the Human Preference Synthetic Dataset (DALLE3 1M+ HQC)~\cite{Egan_Dalle3_1_Million_2024}, which is generated by DALLE3~\cite{betker2023improving} and CogVLM~\cite{wang2024cogvlm}, we sample another 5,000 pairs with challenging prompts and detail-rich images to assess performance in creative and imaginative settings. 
\vspace{-4mm}
\paragraph{Baselines.} We compare our approach with the existing region-based strategy for accelerating diffusion transformer inference.  Region-Adaptive Sampling (RAS) selects regions with high noise magnitude or residual error—often corresponding to areas of high color saturation or brightness. RAS denoises these patches while reusing the output of the last diffusion process for other patches. 

\begin{table}[t]
\centering
% \scriptsize   
\small
\setlength{\tabcolsep}{6pt}
\renewcommand{\arraystretch}{1.1}

\resizebox{\columnwidth}{!}{%
\begin{tabular}{c c c c c c}
\specialrule{1pt}{0.5pt}{0.5pt}
\textbf{Method} 
& \textbf{Steps} 
& \textbf{Ratio} 
& \textbf{time (s)} 
& \textbf{FID} ($\downarrow$) 
& \textbf{CLIP} ($\uparrow$)
\\ 
\specialrule{0.5pt}{0.5pt}{0.5pt}

\textbf{MS-COCO 2017} \\
\specialrule{0.5pt}{0.5pt}{0.5pt}

Original & 10 & 1
& 7.39 & 37.12 & 21.23 \\

\Mname{} & 14 & 0.5
& 6.89 & 35.85 & 21.51 \\

\specialrule{0.5pt}{0.5pt}{0.5pt}

Original & 5 & 1
& 4.09 & 92.19 & 18.91 \\

\Mname{} & 7 & 0.5
& 3.74 & 76.60 & 20.29 \\

\specialrule{0.5pt}{0.5pt}{0.5pt}

\textbf{Dalle3 1M+ HQC} \\
\specialrule{0.5pt}{0.5pt}{0.5pt}

Original & 10 & 1
& 7.31 & 51.48 & 29.66 \\

\Mname{} & 14 & 0.5
& 6.84 & 49.90 & 30.07 \\

\specialrule{0.5pt}{0.5pt}{0.5pt}

Original & 5 & 1
& 4.01 & 107.30 & 23.84 \\

\Mname{} & 7 & 0.5
& 3.69 & 93.95 & 26.22 \\

\specialrule{1pt}{0.5pt}{0.5pt}
\end{tabular}}
\caption{Performance comparison under different steps and sampling ratios.}
\vspace{-6mm}
\end{table}
\label{table2}

\subsection{Efficiency Evaluation}

As shown in Table~\ref{table1}, \Mname{} delivers a stable and predictable trade-off between quality and speed as the sampling ratio decreases. Instead of a sudden degradation, visual fidelity declines smoothly, indicating that the model adapts gracefully to reduced computation. At moderate ratios (around 0.5–0.6), the overall perceptual quality remains close to full sampling, with FID and CLIP scores nearly unchanged and structural metrics (PSNR, SSIM) showing only minor loss.

Even when the sampling ratio is pushed to 0.125, \Mname{} achieves nearly 3× faster inference on the MS-COCO 2017 dataset while preserving coherent structures and consistent semantics. This demonstrates that a large fraction of denoising updates can be safely skipped without compromising perceptual realism.

More importantly, \Mname{} exhibits remarkable fidelity under the Original Image Comparison metrics, showing that its sparse updates remain highly faithful to the original image. The reconstruction error is significantly lower than RAS, and the FID score is reduced by more than half with slightly higher acceleration rates. This suggests that \Mname{} performs sparse sampling in a more structured and semantically aligned way, producing results that are visually closer to full inference while achieving substantially higher efficiency.

% When the sampling ratio decreases, our method exhibits a smooth and well-controlled change in quality rather than an abrupt drop.
% As shown in Table~\ref{table1}, \Mname{} maintains almost full-quality performance when the sampling ratio is moderately reduced.
% At a 0.6 ratio, both FID and CLIP remain close to the full-sampling baseline, and structural metrics such as PSNR and SSIM show only marginal decline, indicating that most perceptual details are preserved.
% Even under the extreme setting of 0.2 ratio, \Mname{} still achieves nearly 2× faster inference (about 1.9× on MS-COCO and 2.0× on DALLE3) while maintaining strong visual coherence and semantic consistency.
% This demonstrates that a large portion of computation can be safely omitted with only a mild and predictable quality loss.

% It is worth noting that in practice, complex or high-frequency regions within an image typically occupy well above 20\% of the total area.
% Thus, even at low sampling ratios, most important regions are still actively updated, ensuring that textures, edges, and semantic boundaries remain sharp and stable across steps.
% In contrast, simply reducing the total number of diffusion steps often causes global degradation and loss of fine details.
% By adjusting the sampling ratio instead, \Mname{} achieves a smoother and more interpretable balance between quality and efficiency, allowing to flexibly select an operating point that suits their computational budget without compromising perceptual realism.

\begin{table*}[h] 
\centering
\small
\setlength{\tabcolsep}{6pt}
\renewcommand{\arraystretch}{1.1}

\resizebox{\textwidth}{!}{
\begin{tabular}{c |c c c c |c c c c c c}
\specialrule{1pt}{0.5pt}{0.5pt}

\multirow{2}{*}{\textbf{Method}} & 
\multirow{2}{*}{\textbf{Steps}} & 
\multirow{2}{*}{\textbf{Ratio}} &
\multirow{2}{*}{\textbf{Time (s)}} & 
\multirow{2}{*}{\textbf{Speedup}} & 
\multicolumn{2}{c}{\textbf{G.T. Comparison}} &
\multicolumn{4}{c}{\textbf{Original Image Comparison}}  \\ [-2pt]
\cmidrule(lr){6-7} \cmidrule(lr){8-11} 
& & & & & 
\textbf{FID} ($\downarrow$) & \textbf{CLIP} ($\uparrow$) &
\textbf{FID} ($\downarrow$) & \textbf{PSNR} ($\uparrow$) & 
\textbf{SSIM} ($\uparrow$) & \textbf{LPIPS} ($\downarrow$) \\
\specialrule{0.5pt}{0.5pt}{0.5pt}

Original & 30 & 1 & 23.72 & - & 29.63 & 21.52 & - & - & - & - \\
\cmidrule[0.5pt]{1-1} 

\multirow{3}{*}{RAS}
& 30 & 0.5 & 13.92 & 1.7× & 32.67 & 21.68 & 11.74 & 22.57 & 0.73 & 0.26 \\
& 30 & 0.25 & 10.31 & 2.3× & 39.25 & 21.78 & 20.82 & 21.23 & 0.68 & 0.24 \\
& 30 & 0.125 & 8.10 & 2.93× & 41.95 & 21.91 & 23.52 & 20.97 & 0.66 & 0.27 \\
\cmidrule[0.5pt]{1-1} 

\multirow{3}{*}{\Mname{}}
& 30 & 0.5 & 14.32 & 1.66× & 30.03 & 21.59 & 4.78 & 25.96 & 0.85 & 0.12 \\
& 30 & 0.25 & 10.29 & 2.31× & 31.62 & 21.65 & 8.18 & 23.37 & 0.77 & 0.13 \\
& 30 & 0.125 & 7.91 & \textbf{3.0×} & 35.50 & 21.84 & 14.23 & 21.85 & 0.71 & 0.20 \\
\specialrule{0.5pt}{1.5pt}{1.5pt}

Original & 10 & 1 & 7.31 & - & 51.48 & 29.66 & - & - & - & - \\
\cmidrule[0.5pt]{1-1} 

\multirow{3}{*}{RAS}
& 10 & 0.65 & 6.03 & 1.22× & 70.34 & 28.49 & 19.26 & 25.03 & 0.76 & 0.17 \\
& 10 & 0.45 & 5.44 & 1.34× & 84.07 & 27.21 & 30.38 & 23.84 & 0.72 & 0.21 \\
& 10 & 0.25 & 4.30 & 1.70× & 104.43 & 25.19 & 49.22 & 23.06 & 0.69 & 0.26 \\
\cmidrule[0.5pt]{1-1} 

\multirow{3}{*}{\Mname{}}
& 10 & 0.6 & 5.80 & 1.26× & 60.41 & 29.06 & 8.78 & 27.72 & 0.84 & 0.09 \\
& 10 & 0.4 & 4.81 & 1.52× & 69.34 & 28.32 & 14.13 & 25.84 & 0.79 & 0.14 \\
& 10 & 0.2 & 3.72 & \textbf{1.97×} & 91.41 & 26.22 & 31.85 & 23.69 & 0.73 & 0.24 \\
\specialrule{0.5pt}{0.5pt}{0.5pt}

\end{tabular}
}
\caption{Performance comparison across different methods on MS-COCO 2017 dataset.}
\end{table*}
\label{table1}

\begin{figure*}[t]
    \centering
    \vspace{-2mm}
\includegraphics[width=0.9\textwidth, trim=10 5 10 5, clip]{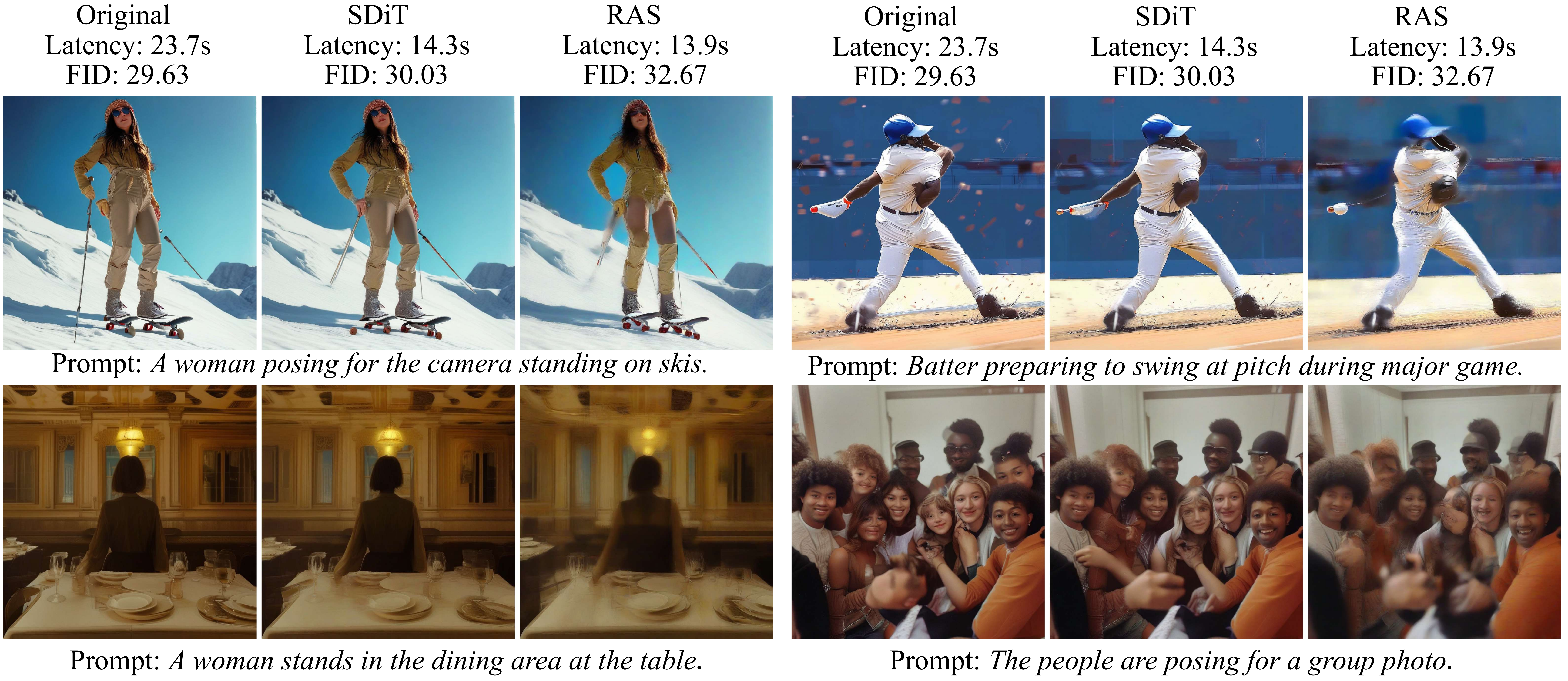}
    \vspace{-2mm}
    \caption{
        Qualitative results. Evaluated by FID against ground truth, \Mname{} reduces inference latency proportionally to the sampling rate, yet preserves fine-grained details and even removes artifacts that appear in full-sampling method.
    }
    \vspace{-4mm}
    \label{fig:sample_image}
\end{figure*}
% In table~\ref{table1}, when the sampling ratio is reduced, our method shows a gradual and well-controlled change in quality rather than an abrupt drop. For example, using 10 denoising steps with only 20\% of the regions updated cuts the total computational cost by about half, yet the resulting images remain highly consistent with the full-sampling baseline. In this setting, FID and sFID increase only slightly, CLIP score stays stable, and PSNR/SSIM indicate strong fidelity to the full-attention output. This means that a large amount of computation can be removed while the quality loss remains small and acceptable.

% Unlike simply reducing the total number of diffusion steps—which often leads to noticeable loss of texture, structure, and semantic accuracy—adjusting the sampling ratio provides a much smoother trade-off between quality and efficiency. Even when the computational budget is very limited, our method still generates images that stay visually coherent, structurally consistent, and faithful to the text prompt. As a result, the drop in quality is gradual and predictable, making it easy to choose an operating point that greatly reduces cost while keeping the output both realistic and semantically accurate.

\vspace{-2mm}
\subsection{Quality Evaluation}

In Table \ref{table2}, our method maintains lower wall-clock time than full-attention sampling while achieving consistently lower FID and visually more accurate results. On the MS-COCO 2017 dataset, \Mname{} attains 1.1× acceleration (4.09 s → 3.74 s) and 15.6 lower FID compared with the baseline. This demonstrates that uniformly updating every patch is not an efficient use of computation.
\vspace{-1mm}
By concentrating denoising on structurally and semantically important regions, our approach utilizes the available budget more effectively, leading to improved perceptual quality and closer alignment with the true data distribution. In short, full attention is no longer the upper bound of quality at a fixed cost—our strategy offers a more principled and effective way to allocate computation in diffusion transformers.

% \paragraph{Noise schedule.}

% \paragraph{Dilate boundaries.}

% TBD
\section{Conclusion}
% In this work, we explored how semantic and structural properties within the latent space of diffusion transformers can be leveraged to achieve efficient and adaptive inference. We found that different regions of an image evolve at highly uneven rates during the denoising process, and that attention patterns tend to cluster along coherent semantic boundaries. Building on these observations, we introduced \Mname{}, a training-free, semantic-region-adaptive framework that dynamically partitions latent representations into coherent regions and allocates computation based on their local complexity. 

% By combining semantic clustering, complexity-aware scheduling, and boundary refinement, \Mname{} achieves up to 3.0× acceleration while maintaining perceptual and structural quality close to full inference and even surpasses uniform baselines under matched budgets.

% Our study highlights that full-attention diffusion is not necessarily the upper bound of quality for a given computational cost. Instead, semantically guided selective denoising provides a more principled path toward efficient, structure-aware generation. We believe this paradigm opens new directions for region-adaptive and context-aware generative modeling, bridging the gap between large-scale diffusion transformers and real-time, deployable visual synthesis.

In this work, we explored how semantic and structural properties within the latent space of diffusion transformers can be leveraged to achieve efficient and adaptive inference.
We found that different regions of an image evolve at highly uneven rates during the denoising process, and that attention patterns tend to cluster along coherent semantic boundaries.
Building on these observations, we introduced \Mname{}, a training-free, semantic region-adaptive framework that dynamically partitions latent representations into coherent regions and allocates computation according to their local complexity.
By combining semantic clustering, complexity-aware scheduling, and boundary refinement, \Mname{} achieves up to 3.0× acceleration while maintaining perceptual and structural quality close to full inference, and surpasses uniform baselines under less computational budgets.
Beyond efficiency, \Mname{} also offers a new perspective on the internal dynamics of diffusion transformers, revealing how spatial semantics and attention structures can be exploited for adaptive computation.
We hope this direction inspires further exploration toward resource-efficient, structure-aware generative modeling.

\newpage
{
    \small
    \bibliographystyle{ieeenat_fullname}
    \bibliography{main}
}

% % WARNING: do not forget to delete the supplementary pages from your submission 
% % \input{sec/X_suppl}

\clearpage
\suppressfloats[t]
\appendix
\section{Additional Ablation Studies}

This appendix provides extended analyses that complement the main paper. 
Unless otherwise stated, all experiments are conducted on the MS-COCO 2017 
dataset at $1024\times1024$ resolution using a Lumina\_Next backbone. 
Sampling steps are fixed to 10 and the sampling ratio to 0.2 by default.

% ============================================================
\subsection{Region Partitioning Ablation}

We first examine how different region-partitioning strategies affect the 
performance of \Mname{}. All variants operate under identical sampling ratios 
and scheduling rules to ensure a fair comparison. As shown in 
Table~\ref{tab:appendix_partition}, semantic clustering–based partitioning 
(Quickshift, SLIC) consistently outperforms naive grid- or patch-based 
alternatives due to better alignment with semantic boundaries.

\begin{table}[t]
\centering
\vspace{4pt}
\caption{Comparison of region partitioning strategies (10 steps, ratio 0.2).}
\label{tab:appendix_partition}
\setlength{\tabcolsep}{6pt}
\renewcommand{\arraystretch}{1.18}
\begin{tabular}{lccc}
\toprule
\textbf{Method} & \textbf{Compute Time} & \textbf{FID}↓ & \textbf{FID\_ori}↓ \\
\midrule
Quickshift      & 0.111  & 83.542   & 51.683  \\
SLIC            & 0.057  & 86.274   & 53.105  \\
K-means         & 0.251  & 108.435  & 81.177  \\
Uniform patch   & 0.378  & 91.981   & 58.740  \\
Direct patch    & 2.251  & 126.250  & 99.418  \\
\bottomrule
\end{tabular}
\end{table}

% ============================================================
\subsection{Complexity Score Ablation}

We further study how different complexity metrics affect the selection of 
top-$k$ regions. As shown in Table~\ref{tab:appendix_complexity_score}, 
our hybrid gradient–noise metric achieves the best performance, indicating 
the importance of capturing both structural and stochastic variations.

\begin{table}[t]
\centering
\vspace{4pt}
\caption{Ablation on complexity score metrics (10 steps, ratio 0.2).}
\label{tab:appendix_complexity_score}
\begin{tabular}{lcc}
\toprule
Complexity Score & FID↓ & FID$_{\text{ori}}$↓ \\
\midrule
$\ell_2$-norm            & 94.503 & 60.912 \\
Noise amplitude          & 96.020 & 62.630 \\
Ours                     & 80.676 & 47.701 \\
Standard deviation       & 94.155 & 60.914 \\
\bottomrule
\end{tabular}
\end{table}

% ============================================================
\subsection{Sensitivity to Sampling Ratio and Dilation}

We evaluate the joint effect of sampling ratio and dilation. 
As seen in Table~\ref{tab:appendix_sampling_ratio}, moderate dilation 
($d=1$ or $d=2$) consistently yields better results, especially under 
aggressive sub-sampling, highlighting the importance of boundary coherence.

\begin{table}[t]
\centering
\vspace{4pt}
\caption{Sensitivity to sampling ratio under different dilation settings.}
\label{tab:appendix_sampling_ratio}
\resizebox{\linewidth}{!}{
\begin{tabular}{cccccc}
\toprule
Dilation & Ratio & Compute (\%) & FID↓ & FID$_{\text{ori}}$↓ \\
\midrule
0 & 0.2 & 15.97 & 94.211 & 61.905 \\
0 & 0.4 & 36.15 & 59.436 & 26.380 \\
0 & 0.6 & 56.36 & 44.963 & 11.741 \\
0 & 0.8 & 77.84 & 39.739 & 6.645 \\
\midrule
1 & 0.2 & 21.30 & 80.676 & 47.701 \\
1 & 0.4 & 43.76 & 50.828 & 17.718 \\
1 & 0.6 & 63.77 & 42.144 & 9.334  \\
1 & 0.8 & 83.04 & 38.549 & 5.297  \\
\midrule
2 & 0.2 & 25.52 & 71.807 & 38.857 \\
2 & 0.4 & 49.45 & 47.149 & 14.484 \\
2 & 0.6 & 69.56 & 40.410 & 7.992  \\
2 & 0.8 & 87.02 & 37.748 & 4.252  \\
\midrule
3 & 0.2 & 29.42 & 65.709 & 32.679 \\
3 & 0.4 & 54.35 & 44.496 & 12.051 \\
3 & 0.6 & 74.49 & 39.486 & 6.865  \\
3 & 0.8 & 89.97 & 37.432 & 3.281  \\
\midrule
full & 1 & 100 & 37.122 & / \\
\bottomrule
\end{tabular}}
\end{table}

% ============================================================
\subsection{Number of Regions $K$}

We vary the quickshift hyper-parameters to change the average number of 
regions. As shown in Table~\ref{tab:avgk_cluster_fid}, increasing region 
granularity helps until segmentation becomes overly fragmented.

\begin{table}[t]
\centering
\vspace{4pt}
\caption{Comparison of clustering configurations.}
\label{tab:avgk_cluster_fid}
\begin{tabular}{c|c|c|c}
\toprule
Average\_K & Cluster & FID↓ & FID$_{\text{ori}}$↓ \\
\midrule
11.650 & 0.164 & 97.884 & 65.136 \\
22.380 & 0.107 & 95.174 & 62.747 \\
30.580 & 0.098 & 94.916 & 62.690 \\
58.510 & 0.097 & 94.913 & 63.043 \\
76.310 & 0.093 & 94.692 & 62.934 \\
\bottomrule
\end{tabular}
\end{table}

% ============================================================
\subsection{Quality Curve Across Sampling Steps}

We evaluate sampling steps $\{3,5,7,10,15,30\}$. As shown in 
Table~\ref{tab:method_step_rate_fid}, \Mname{} provides the largest gains 
at low-step regimes, where spatial complexity is highly uneven. 
As steps increase, the denoising trajectory becomes smoother and the 
advantage naturally narrows.

\begin{table}[h]
\centering
\vspace{4pt}
\caption{Quality comparison across sampling steps.}
\label{tab:method_step_rate_fid}
\begin{tabular}{c|c|c|c|c}
\toprule
Method & Step & Rate & Time & FID↓ \\
\midrule
Full & 3  & 1   & 2.436   & 248.653 \\
Full & 5  & 1   & 3.824   & 92.192  \\
Full & 7  & 1   & 5.280   & 51.753  \\
Full & 10 & 1   & 7.393   & 37.122  \\
Full & 15 & 1   & 10.962  & 31.512  \\
Full & 30 & 1   & 23.724  & 29.627  \\
\midrule
\Mname{} & 5  & 0.3 & 2.359   & 127.136 \\
\Mname{} & 7  & 0.5 & 3.721   & 76.601  \\
\Mname{} & 10 & 0.5 & 5.208   & 44.882  \\
\Mname{} & 14 & 0.5 & 6.885   & 35.855  \\
\Mname{} & 20 & 0.6 & 11.036  & 31.026  \\
\Mname{} & 40 & 0.6 & 21.805  & 29.814  \\
\bottomrule
\end{tabular}
\end{table}

% ============================================================
\section{Additional Qualitative Results}

We provide qualitative comparisons under different sampling ratios and 
sampling steps.

% -------------------- FIGURE: 10 STEPS --------------------
\begin{figure*}[t]
\centering
\vspace{4pt}
\includegraphics[width=\linewidth]{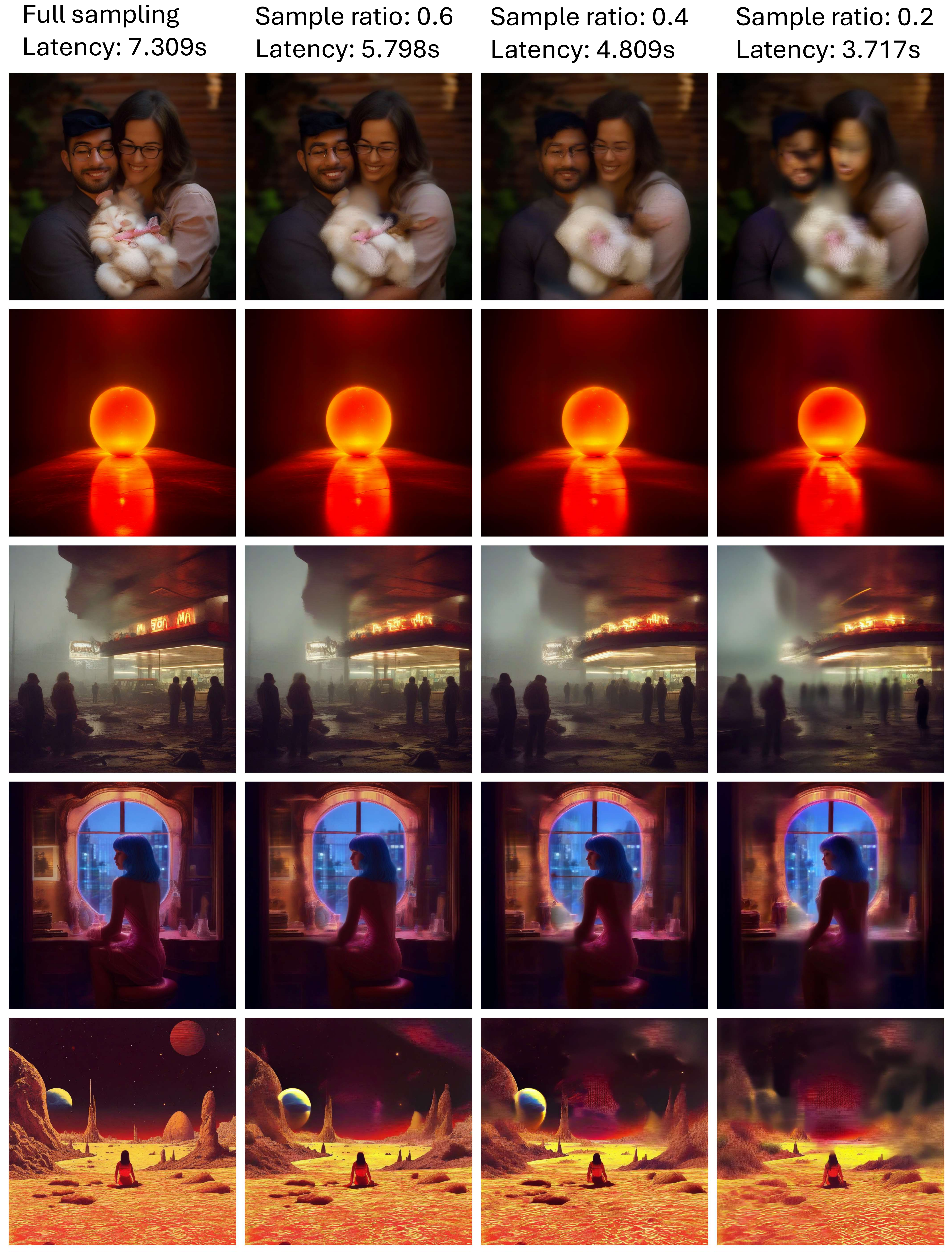}
\caption{\textbf{Qualitative comparison under 10 sampling steps.}}
\label{fig:qual_10steps}
\end{figure*}

% -------------------- FIGURE: 30 STEPS --------------------
\begin{figure*}[t]
\centering
\vspace{4pt}
\includegraphics[width=\linewidth]{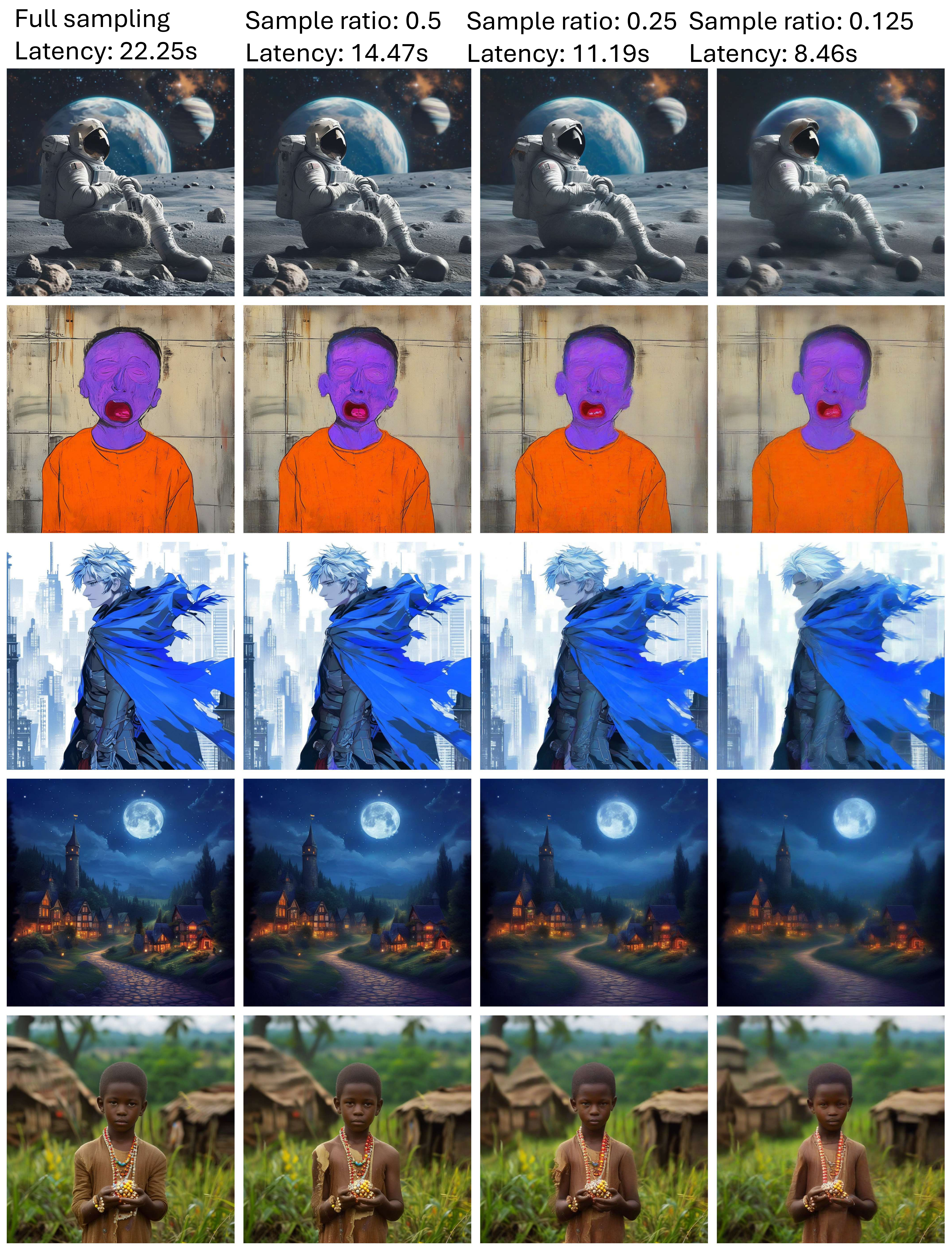}
\caption{\textbf{Qualitative comparison under 30 sampling steps.}}
\label{fig:qual_30steps}
\end{figure*}

\vspace{10pt}

We further include complete quantitative comparisons on 
MS-COCO 2017 and DALLE3 1M+ HQC datasets for both 30-step and 10-step settings.

% ============================================================
\begin{table*}[h]
\centering
\caption{Performance comparison on MS-COCO 2017 under step=30 and step=10.}
\label{tab:coco_results}
\resizebox{\linewidth}{!}{
\begin{tabular}{c|c|c|c|c|c|c|c|c|c}
\toprule
Method & Step & Rate & Cluster & Time & FID↓ & FID\_ori↓ & PSNR↑ & SSIM↑ & LPIPS↓ \\
\midrule
\multicolumn{10}{c}{\textbf{Step = 30}} \\
\midrule
Lumina\_Next & 30 & 1 & / & 23.724 & 29.627 & / & / & / & / \\
RAS          & 30 & 0.5 & / & 13.924 & 32.666 & 11.738 & 22.566 & 0.732 & 0.260 \\
\Mname{}     & 30 & 0.5 & 0.363 & 14.324 & 30.031 & 4.775 & 25.957 & 0.850 & 0.120 \\
RAS          & 30 & 0.25 & / & 10.308 & 39.248 & 20.816 & 21.226 & 0.677 & 0.243 \\
\Mname{}     & 30 & 0.25 & 0.348 & 10.293 & 31.617 & 8.184 & 23.366 & 0.774 & 0.125 \\
RAS          & 30 & 0.125 & / & 8.095 & 41.945 & 23.522 & 20.973 & 0.661 & 0.270 \\
\Mname{}     & 30 & 0.125 & 0.366 & 7.907 & 35.500 & 14.228 & 21.851 & 0.709 & 0.196 \\
\midrule
\multicolumn{10}{c}{\textbf{Step = 10}} \\
\midrule
Lumina\_Next & 10 & 1 & / & 7.393 & 37.122 & / & / & / & / \\
RAS          & 10 & 0.65 & / & 6.170 & 54.340 & 21.269 & 24.404 & 0.742 & 0.184 \\
\Mname{}     & 10 & 0.6 & 0.104 & 5.615 & 42.562 & 8.788 & 24.969 & 0.773 & 0.157 \\
RAS          & 10 & 0.45 & / & 5.453 & 70.265 & 36.854 & 23.297 & 0.705 & 0.235 \\
\Mname{}     & 10 & 0.4 & 0.112 & 4.892 & 49.639 & 13.705 & 25.214 & 0.775 & 0.148 \\
RAS          & 10 & 0.25 & / & 4.295 & 93.314 & 60.050 & 22.605 & 0.677 & 0.284 \\
\Mname{}     & 10 & 0.2 & 0.114 & 3.821 & 73.800 & 35.137 & 23.258 & 0.710 & 0.247 \\
\bottomrule
\end{tabular}}
\end{table*}

\vspace{10pt}

% ============================================================
\begin{table*}[t]
\centering
\caption{Performance comparison on DALLE3 1M+ HQC dataset under step=30 and step=10.}
\label{tab:dalle_results}
\resizebox{\linewidth}{!}{
\begin{tabular}{c|c|c|c|c|c|c|c|c|c}
\toprule
Method & Step & Rate & Cluster & Time & FID↓ & FID\_ori↓ & PSNR↑ & SSIM↑ & LPIPS↓ \\
\midrule
\multicolumn{10}{c}{\textbf{Step = 30}} \\
\midrule
Lumina\_Next & 30 & 1 & / & 22.253 & 27.514 & / & / & / & / \\
RAS          & 30 & 0.5 & / & 13.904 & 41.725 & 15.575 & 22.603 & 0.724 & 0.167 \\
\Mname{}     & 30 & 0.5 & 0.341 & 14.466 & 31.088 & 5.586 & 26.242 & 0.852 & 0.062 \\
RAS          & 30 & 0.25 & / & 10.288 & 50.783 & 22.810 & 21.541 & 0.679 & 0.228 \\
\Mname{}     & 30 & 0.25 & 0.309 & 11.185 & 37.129 & 10.578 & 23.529 & 0.772 & 0.126 \\
RAS          & 30 & 0.125 & / & 8.100 & 59.052 & 29.850 & 20.915 & 0.649 & 0.281 \\
\Mname{}     & 30 & 0.125 & 0.358 & 8.456 & 47.269 & 18.786 & 21.889 & 0.703 & 0.202 \\
\midrule
\multicolumn{10}{c}{\textbf{Step = 10}} \\
\midrule
Lumina\_Next & 10 & 1 & / & 7.309 & 51.480 & / & / & / & / \\
RAS          & 10 & 0.65 & / & 6.032 & 70.339 & 19.259 & 25.028 & 0.760 & 0.165 \\
\Mname{}     & 10 & 0.6 & 0.109 & 5.798 & 60.411 & 8.781 & 27.718 & 0.840 & 0.090 \\
RAS          & 10 & 0.45 & / & 5.442 & 84.073 & 30.377 & 23.843 & 0.724 & 0.214 \\
\Mname{}     & 10 & 0.4 & 0.104 & 4.809 & 69.340 & 14.134 & 25.845 & 0.794 & 0.140 \\
RAS          & 10 & 0.25 & / & 4.301 & 104.433 & 49.221 & 23.058 & 0.694 & 0.263 \\
\Mname{}     & 10 & 0.2 & 0.105 & 3.717 & 91.414 & 31.850 & 23.693 & 0.727 & 0.240 \\
\bottomrule
\end{tabular}}
\end{table*}

\end{document}